\documentclass{article}

\usepackage{subcaption}
\usepackage{graphicx}
\usepackage{booktabs}
\usepackage{wrapfig}
\usepackage[table]{xcolor}
\usepackage{amsmath}
\usepackage{algorithm}
\usepackage{amsthm}
\usepackage{algorithmic}

\usepackage{makecell}
\usepackage{threeparttable}
\usepackage{multirow}
\newcommand{\up}{\ensuremath{\uparrow}}
\newcommand{\down}{\ensuremath{\downarrow}}

 \usepackage[preprint]{neurips_2026}

\usepackage[utf8]{inputenc} 
\usepackage[T1]{fontenc}    
\usepackage{hyperref}       
\usepackage{url}            
\usepackage{booktabs}       
\usepackage{amsfonts}       
\usepackage{nicefrac}       
\usepackage{microtype}      
\usepackage{xcolor}         
\definecolor{bestgray}{gray}{0.92}
\definecolor{mygreen}{RGB}{0,128,0}
\definecolor{myred}{RGB}{190,30,45}
\definecolor{mygray}{gray}{0.45}
\hypersetup{
    colorlinks=true,
}
\renewcommand{\up}{\textcolor{mygreen}{$\uparrow$}}
\renewcommand{\down}{\textcolor{myred}{$\downarrow$}}

\title{Temporal Concentration from Rollout Errors: Implicit Preference Optimization for Text-to-Video Generation}

\author{
Henglin Liu$^{1,2*}$,
Fangyuan Kong$^{2,\S}$,
Jing Wang$^{2,3*}$,
Yizhou Lin$^{1}$, \\
\bfseries
Nisha Huang$^{1}$,
Chang Liu$^{1}$,
Xintao Wang$^{2}$,
Pengfei Wan$^{2}$,
Kun Gai$^{2}$,
Xiu Li$^{1,\dagger}$\\[0.5em]
\normalfont
$^{1}$Tsinghua University\\
$^{2}$Kling Team, Kuaishou Technology\\
$^{3}$Sun Yat-sen University\\[0.5em]
\textit{\href{https://henglin-liu.github.io/cIPO_vis/}{Project Page}}
\quad
$^{\S}$Project Leader.\quad
$^{\dagger}$Corresponding Author.\quad
$^{*}$Work Conducted During Internship.\\[0.5em]
\texttt{liu-hl24@mails.tsinghua.edu.cn}
}

\begin{document}

\maketitle

\begin{abstract}

Recent advances in preference alignment for diffusion-based video generation, particularly via Direct Preference Optimization (DPO), have significantly improved visual quality. However, temporally sparse artifacts such as motion collapse, object flickering, and color oversaturation remain a major barrier to perceptual realism. Existing methods struggle with these issues due to two key limitations: (1) the preference attribution bottleneck, where offline human annotations are costly and fail to accurately capture learning dynamics, while online reward signals are rollout-aware but often unstable and biased; and (2) temporal credit misallocation, where uniformly applied supervision cannot effectively target the brief segments in which artifacts occur. To address these challenges, we propose concentrated Implicit Preference Optimization (cIPO), a post-training framework for video diffusion models. cIPO derives implicit preference signals directly from the denoising process: given a real video, the model adds forward noise and reconstructs it via iterative denoising, treating the original as the preferred sample and the reconstruction as the dispreferred one. This formulation captures inference-time errors without requiring human annotations or external reward models. Moreover, frame-level discrepancies between original and reconstructed videos reveal when failures occur. cIPO leverages this by computing temporal reconstruction errors and concentrating optimization on high-error segments, enabling more precise correction of failure-prone regions. Extensive experiments demonstrate that cIPO consistently enhances video authenticity and temporal coherence across multiple datasets, highlighting the effectiveness and efficiency of implicit preference with temporally concentrated optimization.
\end{abstract}

\section{Introduction}
\begin{figure}
  \centering
  \includegraphics[width=\linewidth]{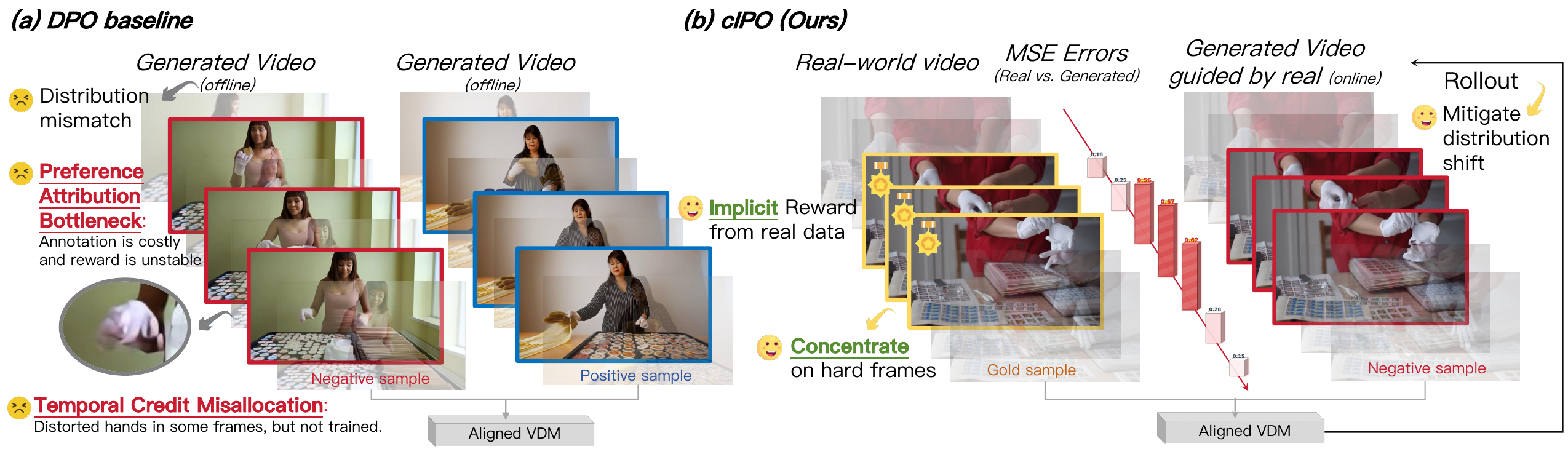}
  \caption{Compared with the baseline method, our method does not rely on additional human annotations or reward models. Besides, it can automatically identify difficult frames for training.}
  \label{fig:teaser}
  \vspace{-12pt}
\end{figure}
Recent advances in diffusion models have substantially improved the visual quality. However, generating temporally coherent and artifact-free videos remains challenging. Unlike image generation, where perceptual failures are often spatially localized, video generation failures are often sparse in time: a video may appear plausible for most frames while exhibiting severe artifacts in only a few short segments, such as motion collapse, object flickering, or abrupt temporal discontinuities. These localized failures significantly degrade perceived video quality.

Preference optimization has recently emerged as a promising paradigm for post-training. By optimizing relative preferences between preferred and dispreferred outputs, methods such as direct preference optimization (DPO) offer an effective alternative to explicit reward modeling. However, applying preference optimization to video generation presents two key challenges: preference attribution bottleneck and temporal credit misallocation.

\begin{wrapfigure}{r}{0.3\textwidth}
  \centering
  \includegraphics[width=\linewidth]{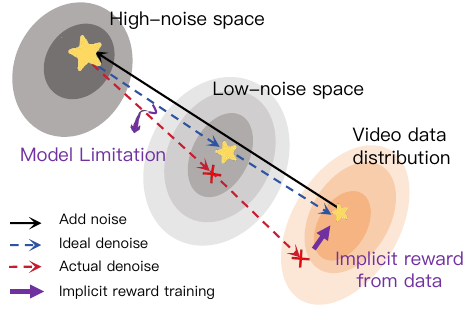}
  \caption{Ideal denoising should restore the video, but model trajectories drift due to error accumulation, naturally defining an implicit training preference.}
  \label{modelsize_denoise}
\vspace{-10pt} 
\end{wrapfigure}
The first challenge is the preference attribution bottleneck (\textit{where reliable preference signals should come from}). As illustrated in Fig.~\ref{modelsize_denoise}, the core difficulty in video diffusion is not merely generating samples, but faithfully following the ideal reverse denoising trajectory back to the video data manifold~\cite{songscore,lipmanflow}. Starting from a perturbed real video (black arrow), an ideal denoising process should progressively recover the original sample and return to the data distribution (blue dashed arrow). In practice, however, due to limitations in model generative capacity, the learned denoising trajectory often deviates from this ideal path and accumulates errors over iterative rollout (red dashed arrow), leading to perceptual degradation in the final generated video. This rollout-induced deviation exposes a fundamental limitation of existing preference supervision. A straightforward solution is to collect human preference annotations by directly comparing generated videos. While such supervision is often high-quality, it is prohibitively expensive to scale and too static to track the model’s evolving denoising behavior online. An alternative is to train reward models to score videos generated by the current policy. This captures rollout behavior but offers only indirect scalar signal on final outputs, and relies on external reward models, whose inherent biases may misalign with perceptual quality and induce optimization instability and reward hacking. 
The key issue is that preference signals should be both rollout-aware, so that they reflect the model’s actual iterative generation behavior, and self-grounded, so that they arise from the model’s own denoising dynamics rather than an external evaluator. This suggests a simple alternative paradigm: directly treat the discrepancy between the ideal target (real video) and the model’s actual denoised result as an implicit preference signal, and optimize the model to reduce this rollout-induced drift (as shown by the purple arrows in Fig.~\ref{modelsize_denoise}).

\begin{wrapfigure}{r}{0.3\textwidth}
\vspace{-12pt} 
  \centering
  \includegraphics[width=\linewidth]{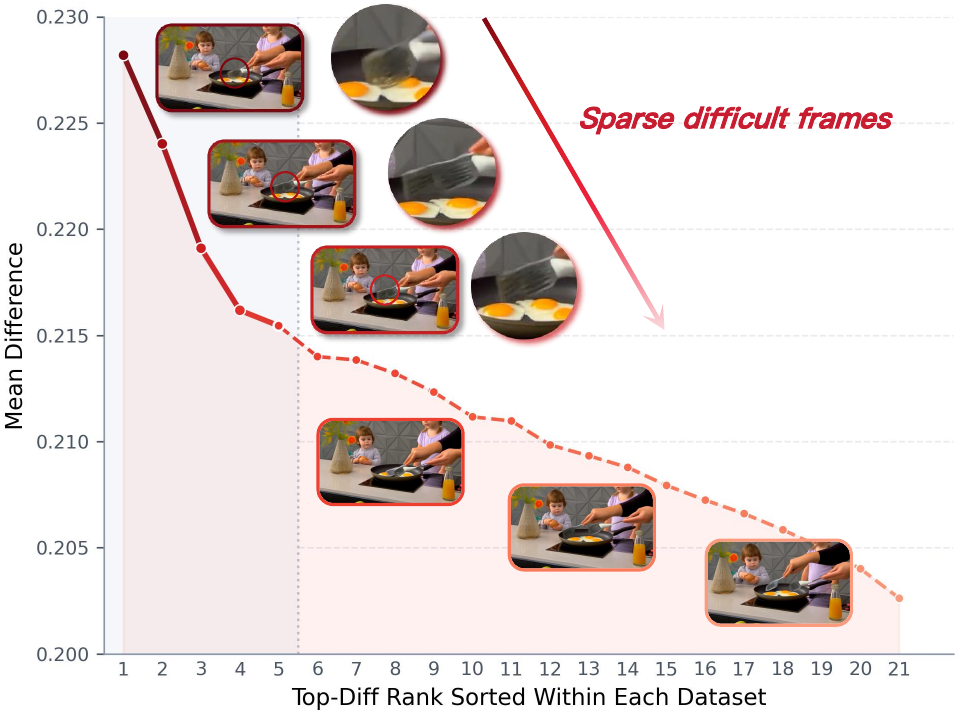}
  \caption{A few frames exhibit significantly larger errors.}
  \label{dataset_topdiff}
\vspace{-12pt} 
\end{wrapfigure}
The second challenge is temporal credit misallocation (\textit{where optimization should be applied}). Even with reliable preference pairs, video preference optimization remains inefficient if optimization is distributed uniformly over time. We observe that errors in video diffusion are highly non-uniform (as shown in Fig.~\ref{dataset_topdiff}): A few frames exhibit significantly larger errors. Under such sparse failure patterns, uniform supervision wastes optimization on already well-generated frames while weakening learning on the temporally localized segments that matter most.

To address these problems, we propose concentrated Implicit Preference Optimization (cIPO), a simple and effective framework for aligning video diffusion models. For preference attribution bottleneck, we constructs implicit preference supervision directly from diffusion denoising process. Given a real video, cIPO first perturbs it with forward diffusion noise, then reconstructs it through iterative denoising using the policy model online. This reconstruction process explicitly rolls out the model’s own inference-time denoising trajectory, exposing the accumulated errors that emerge only during multi-step generation. The discrepancy between the original video and its denoising result yields a naturally aligned preference signal: the original video serves as the preferred sample, while the reconstructed video serves as the dispreferred one. This formulation naturally aligns with the model’s actual inference process, mitigating the training–inference mismatch without requiring human preference labels or external reward models.

To address temporal credit misallocation, cIPO further introduces concentrated preference optimization. Specifically, cIPO computes temporal reconstruction discrepancies, aggregates them over temporal windows, and allocates preference supervision only to the highest-error segments. By concentrating optimization on the temporal regions most responsible for perceptual degradation, cIPO substantially improves the density, precision, and effectiveness of preference optimizations.

Our work makes three contributions:
(1) We identify preference attribution and temporal credit assignment as two central bottlenecks in video preference learning, demonstrating that video generation significantly exacerbates these challenges due to its iterative denoising trajectory and temporally sparse failure patterns. (2) We propose cIPO, a post-training framework that combines implicit denoised-based preference with concentrated temporal optimization. (3) We demonstrate that cIPO consistently improves video authenticity and temporal coherence across multiple benchmarks, showing that video preference learning benefits from implicit preference and concentrated temporal optimization.

\section{Related Work}


\subsection{Preference Learning for Video Generation}
Video diffusion preference optimization methods can be categorized by supervision source: human annotations, reward models, or synthetic data. Human-annotated methods directly optimize against pairwise labels. Flow-DPO\cite{liuimproving} adapts DPO to video, while DenseDPO\cite{wudensedpo} uses dense segment-level preferences to reduce motion bias. These rely on costly offline annotations and are weakly aligned with current model failures. Reward-based methods use trained reward models. \cite{liuimproving,wang2024lift,wang2024lift,wangunified,zhang2024onlinevpo} score outputs for RL optimization. While these scale well and offer rollout-aware feedback, they are limited by training instability and reward hacking. Synthetic data methods avoid annotation by heuristically degrading real videos. DF-DPO\cite{cheng2025discriminator} uses heuristic degradations (e.g., temporal reversal, frame shuffling) as negatives, LocalDPO\cite{huang2026mind} enhances this with localized corruptions, and RealDPO~\cite{cheng2025realdpo} directly contrasts real videos with model-generated outputs offline. However, it still does not explicitly capture how errors emerge during the denoising process itself. Overall, existing methods depend on costly or biased external reward, whereas cIPO leverages internal online denoising dynamics for precise, adaptive preferences.

\subsection{Dynamic Temporal Sampling in Video Generation}
Recent studies on dynamic sampling in video generation exploit temporal non-uniformity to achieve non-uniform computation allocation along the temporal dimension. A representative line of work includes DLFR-VAE~\cite{yuan2025dlfr}, VGDFR~\cite{yuan2025vgdfr}, and DLFR-Gen~\cite{Yuan_2025_IccV}, which observe that motion and information density vary substantially across time, and thus dynamically allocate fewer latent tokens to low-motion segments while preserving denser temporal representations for high-motion regions. Concretely, DLFR-VAE~\cite{yuan2025dlfr} introduces a dynamic latent frame-rate scheduler in the VAE latent space, while VGDFR~\cite{yuan2025vgdfr} and DLFR-Gen~\cite{Yuan_2025_IccV} extend this idea to diffusion-based video generation via adaptive latent frame-rate scheduling and latent frame merging during denoising.
Existing work dynamically redistributes computation for efficient generation, whereas our method dynamically redistributes alignment signal for temporally precise preference optimization, directly targeting sparse temporal artifacts such as flickering, motion collapse, and abrupt discontinuities.

\section{Method}
\begin{figure}
  \centering
  \includegraphics[width=\linewidth]{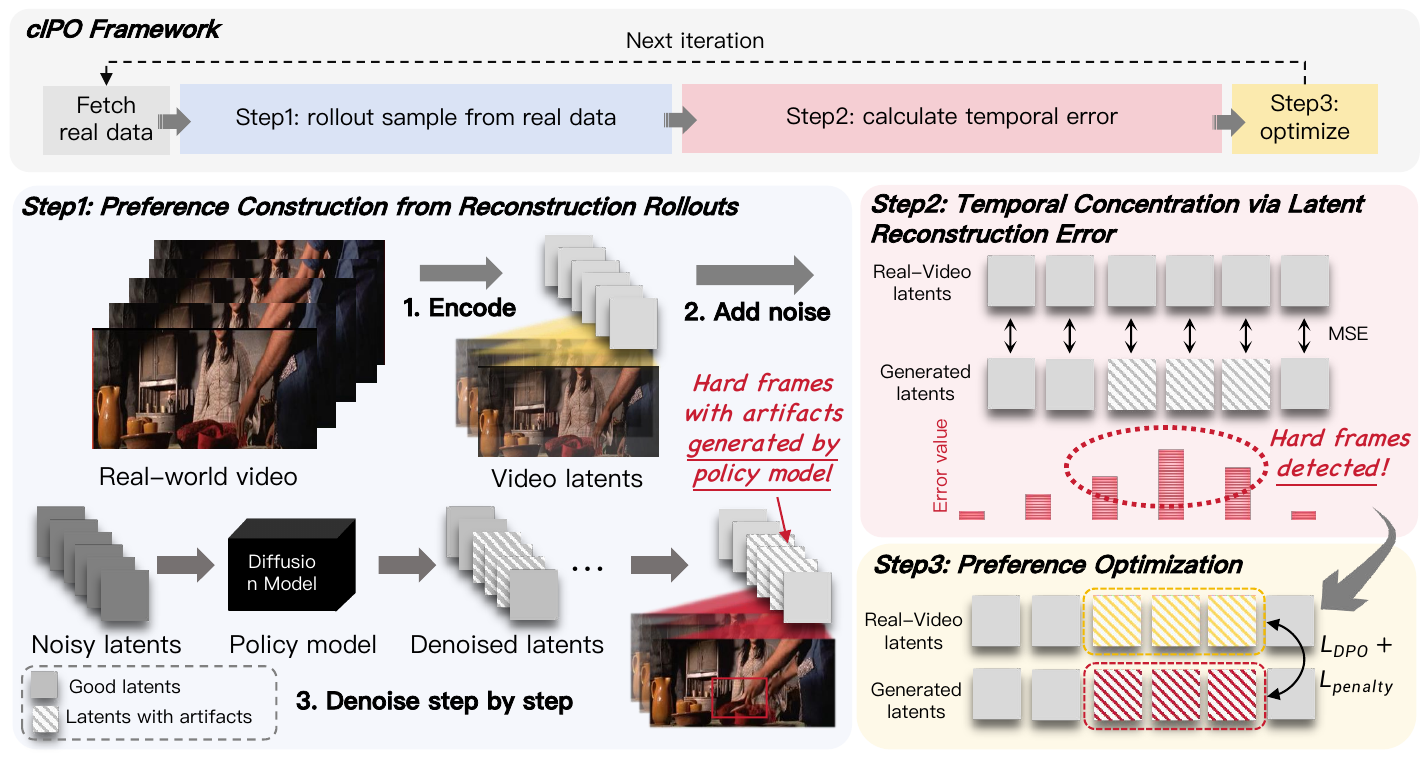}
  \caption{The figure shows the three-stage pipeline of cIPO. It first builds implicit preference pairs by reconstructing real videos through diffusion rollouts, using the original video as the preferred sample and its reconstruction as the dispreferred one. It then detects temporally localized hard frames via frame-wise latent reconstruction error, and finally concentrates DPO optimization on these high-error segments to improve temporal coherence and visual quality. Note that the entire process is carried out in the latent space. The videos in the figure are only for visual demonstration.}
  \label{method}
  \vspace{-12pt}
\end{figure}

\label{sec:method}

As shown in Fig.~\ref{method}, cIPO framework consists of three components: (1) an implicit preference construction mechanism that derives preference pairs from reconstruction rollouts without external rewards; (2) a temporally concentrated selection strategy that focuses optimization on failure-prone segments; and (3) a pairwise preference objective that improves temporal fidelity while preserving already-correct content.

\subsection{Implicit Preference Construction from Reconstruction Rollouts}
\label{subsec:implicit_pref}
Let $x \in \mathcal{X}$ denote a real video sampled from the training distribution, and let $c \in \mathcal{C}$ denote its text condition. We encode $x$ into the latent space of a pretrained video VAE encoder $\mathcal{E}$:
\begin{equation}
z_0 = \mathcal{E}(x) \in \mathbb{R}^{C \times T \times H \times W},
\label{eq:latent_clean}
\end{equation}
where $T$ is the number of latent frames. Let $f_\theta$ denote the trainable video diffusion model and $f_{\theta_{\mathrm{ref}}}$ a frozen reference model. Given a diffusion starting index $s$, we construct a partially corrupted latent by applying the forward diffusion process to $z_0$. Specifically, let $\{t_k\}_{k=0}^{N-1}$ denote the scheduler timesteps and let $\epsilon \sim \mathcal{N}(0, I)$. We form $z_{t_s} = q(z_{t_s} \mid z_0, \epsilon)$, where $q$ is the scheduler-induced forward noising process at timestep $t_s$. Starting from $z_{t_s}$, we run the reverse diffusion process conditioned on $c$ using $f_\theta$ over the remaining timesteps. Denoting one reverse update by $\Phi_\theta$, we iteratively compute
\begin{equation}
z_{t_{k+1}} = \Phi_\theta(z_{t_k}, c, t_k),
\qquad k = s, \dots, N-1,
\label{eq:reverse_step}
\end{equation}
and denote the terminal denoised latent by $\tilde{z}_0^{(s)}.$. We then define an implicit preference pair
\begin{equation}
y^{+} = z_0,
\qquad
y^{-} = \tilde{z}_0^{(s)},
\label{eq:implicit_pair}
\end{equation}
where the clean latent is treated as the preferred sample and its noised-then-denoised reconstruction as the dispreferred sample. This yields one preference pair per training instance without requiring human annotations or a learned reward model. The specific algorithm process can refer to ~\ref{app_rollout}.

We found that the reconstruction effect of the first frame is very poor. To reduce the first-frame drift and video structure collapse, we introduce a \emph{clean first-frame anchoring} strategy. After obtaining $y^{-}$, we replace its first latent frame with that of the clean sample:
\begin{equation}
\hat{y}_{\tau}^{-} =
\begin{cases}
y^{+}_{\tau}, & \tau = 1,\\
y^{-}_{\tau}, & \tau > 1.
\end{cases}
\label{eq:first_frame_anchor}
\end{equation}
The anchored negative $\hat{y}^{-}$ preserves a stable reference for appearance and scene layout, while allowing later frames to expose temporal inconsistencies.

\subsection{Temporal Concentration via Latent Reconstruction Error}
\label{subsec:temporal_concentration}

A key observation is that temporal artifacts in generated videos are often concentrated in a few short segments. If preference optimization is applied uniformly over all frames, the gradients contributed by normal frames can overwhelm those from genuinely problematic segments, causing optimization to overfit temporally uninformative regions. cIPO addresses this issue by identifying the highest-error contiguous temporal window and restricting preference learning to that window.

For a preferred sample $y^{+}$ and an anchored negative $\hat{y}^{-}$, we define a framewise latent reconstruction discrepancy
\begin{equation}
d_{\tau}
=
\frac{1}{CHW}
\left\|
y^{+}_{\tau} - \hat{y}^{-}_{\tau}
\right\|_2^2,
\qquad
\tau = 1,\dots,T.
\label{eq:framewise_discrepancy}
\end{equation}
Here $d_{\tau}$ measures how severely the reconstruction deviates from the clean video at frame $\tau$, $T$ is the total number of frames in the original video. Given a window length $K$, we score each contiguous temporal window by its average discrepancy:
\begin{equation}
\mathcal{D}(a)
=
\frac{1}{K}
\sum_{\tau=a}^{a+K-1}
d_{\tau},
\qquad
a \in \{1,\dots,T-K+1\}.
\label{eq:window_score}
\end{equation}
We then select the most failure-prone window
\begin{equation}
a^{\star}
=
\arg\max_{a \in \{1,\dots,T-K+1\}}
\mathcal{D}(a),
\qquad
W^{\star}
=
\{a^{\star}, a^{\star}+1, \dots, a^{\star}+K-1\}.
\label{eq:best_window}
\end{equation}
Finally, we restrict both the preferred and dispreferred samples to the selected window:
\begin{equation}
\bar{y}^{+} = y^{+}_{W^{\star}},
\qquad
\bar{y}^{-} = \hat{y}^{-}_{W^{\star}}.
\label{eq:concentrated_pair}
\end{equation}

The use of a \emph{contiguous} window is essential. Selecting scattered high-error frames independently may destroy short-range temporal structure and produce an incoherent optimization target that breaks short-range motion dependencies and weakens temporal consistency across adjacent frames. 
By contrast, Eq.~\eqref{eq:best_window} preserves local motion continuity and concentrates supervision on a semantically meaningful segment of the video.



\subsection{Concentrated Implicit Preference Optimization}
\label{subsec:cIPO_objective}

We now define the optimization objective. Let $\ell_{\theta}(y; c, t, \epsilon)$ denote the per-sample diffusion reconstruction loss of model $f_\theta$ on a latent video $y$, evaluated at diffusion step $t$ with noise realization $\epsilon$:
\begin{equation}
\ell_{\theta}(y; c, t, \epsilon)
=
\frac{1}{|\Omega|}
\left\|
f_{\theta}(y_t, c, t) - u(y, t, \epsilon)
\right\|_2^2,
\label{eq:diffusion_loss}
\end{equation}
where $u(y,t,\epsilon)$ is the scheduler-specific regression target~\cite{liuimproving} and $|\Omega|$ is the number of latent elements. For a concentrated preferred sample $\bar{y}^{+}$ and a concentrated negative $\bar{y}^{-}$, we define the model-reference reconstruction gap
\begin{equation}
\Delta_{\theta}(y)
=
\ell_{\theta}(y; c, t, \epsilon)
-
\ell_{\theta_{\mathrm{ref}}}(y; c, t, \epsilon).
\label{eq:model_ref_gap}
\end{equation}
Intuitively, $\Delta_{\theta}(y)$ measures whether the current model improves upon or degrades relative to the reference model on sample $y$. Given a concentrated preferred sample $\bar{y}^{+}$ and its corresponding concentrated negative $\bar{y}^{-}$, we define a pairwise preference logit
\begin{equation}
g
=
-\frac{\beta}{2}
\left[
\Delta_{\theta}(\bar{y}^{+})
-
\Delta_{\theta}(\bar{y}^{-})
+
\lambda \, \mathrm{ReLU}\!\left(\Delta_{\theta}(\bar{y}^{+})\right)
\right].
\label{eq:pair_logit}
\end{equation}
where $\beta > 0$ is the preference sharpness coefficient and $\lambda \ge 0$ controls a stability regularizer.

The first term in Eq.~\eqref{eq:pair_logit} encourages the trainable model to achieve a larger relative improvement on the preferred sample than on the dispreferred one. The second term,
\(
\mathrm{ReLU}(\Delta_{\theta}(\bar{y}^{+})),
\)
acts as a \emph{winner-preservation penalty} following Smaug~\cite{pal2024smaug}: if the current model performs worse than the reference on the preferred clean segment, the penalty activates and suppresses this undesirable drift. This term is particularly important in the video setting, where aggressive preference updates can otherwise damage already-correct frames (see details in~\ref{sec:flowdpo_gradient_degenerate}).

The final cIPO objective adopts the standard DPO-style pairwise preference form $\mathcal{L}_{\mathrm{cIPO}}
=
-\mathbb{E}_{(x,c)}
\left[
\log \sigma\!\left(-g\right)
\right]$
, where $\sigma(\cdot)$ is the sigmoid function. This objective instantiates the standard DPO negative log-sigmoid loss with reconstruction-induced implicit preferences, encouraging the model to assign lower relative reconstruction error to the preferred clean segment than to its corresponding hard negative.

\section{Experiment}
\begin{table*}[t]
\centering
\small
\setlength{\tabcolsep}{4.5pt}
\renewcommand{\arraystretch}{1.12}
\begin{threeparttable}
\caption{
Quantitative Comparison on MotionBench prompts from authenticity and motion dimensions. Best results are highlighted in \textbf{bold}, and second-best results are underlined. Forensic is the abbreviation of Forensic-chat, Om-D is the abbreviation of OmniAID-Dino, Off/On is the abbreviation of offline/online, and GT is the abbreviation of ground truth (i.e., real-world video).
}
\label{main_exp}
\begin{tabular}{lccccccc}
\toprule
\textbf{Method}
& \multicolumn{2}{c}{\textbf{Frame Authenticity}}
& \multicolumn{5}{c}{\textbf{Temporal Quality (Vbench)}} \\
\cmidrule(lr){2-3} \cmidrule(lr){4-8}
& \textbf{Forensic} $\uparrow$
& \textbf{Om-D} $\uparrow$
& \textbf{Background} $\uparrow$
& \textbf{Motion} $\uparrow$
& \textbf{Subject} $\uparrow$
& \textbf{Temporal} $\uparrow$
& \textbf{Overall} $\uparrow$ \\
\midrule

Pretrained
& 0.804 & 0.479 & 0.937 & 0.974 & 0.929 & 0.960 & 0.161 \\

DPO (Off)
& 0.810 & 0.474 & 0.938 & 0.975 & 0.927 & 0.960 & 0.161 \\

DPO (Off, GT)
& 0.815 & 0.478 & \underline{0.939} & 0.972 & 0.919 & 0.957 & 0.162 \\

DPO (On, GT)
& 0.830 & \underline{0.486} & 0.931 & 0.973 & 0.919 & 0.958 & \underline{0.162} \\

DenseDPO
& \underline{0.819} & 0.475 & 0.937 & \underline{0.975} & \underline{0.930} & \textbf{0.963} & 0.162 \\

\rowcolor{bestgray}
Ours
& \textbf{0.876}
& \textbf{0.524}
& \textbf{0.947}
& \textbf{0.989}
& \textbf{0.937}
& \underline{0.961}
& \textbf{0.168} \\
\bottomrule
\end{tabular}
\end{threeparttable}
\end{table*}

\begin{table*}[t]
\centering
\small
\setlength{\tabcolsep}{4.5pt}
\renewcommand{\arraystretch}{1.12}
\begin{threeparttable}
\caption{
Quantitative Comparison on WISA prompts from authenticity and motion dimensions. Best results are highlighted in \textbf{bold}, and second-best results are underlined.
}
\label{main_exp_wisa_normal}
\begin{tabular}{lccccccc}
\toprule
\textbf{Method}
& \multicolumn{2}{c}{\textbf{Frame Authenticity}}
& \multicolumn{5}{c}{\textbf{Temporal Quality (VBench)}} \\
\cmidrule(lr){2-3} \cmidrule(lr){4-8}
& \textbf{Forensic} $\uparrow$
& \textbf{Om-D} $\uparrow$
& \textbf{Background} $\uparrow$
& \textbf{Motion} $\uparrow$
& \textbf{Subject} $\uparrow$
& \textbf{Temporal} $\uparrow$
& \textbf{Overall} $\uparrow$ \\
\midrule

Pre-trained
& 0.909 & 0.486 & 0.951 & 0.985 & 0.956 & 0.975 & 0.217 \\

DPO (Off)
& 0.908 & 0.504 & 0.949 & 0.987 & 0.961 & 0.978 & 0.225 \\

DPO (Off, GT)
& 0.916 & 0.491 & \textbf{0.955} & 0.985 & 0.960 & 0.976 & 0.222 \\

DPO (On, GT)
& \underline{0.917} & \underline{0.504} & 0.951 & 0.987 & 0.961 & 0.979 & \underline{0.226} \\

DenseDPO
& 0.911 & 0.504 & 0.950 & \underline{0.987} & \underline{0.962} & \underline{0.979} & 0.223 \\

\rowcolor{bestgray}
Ours
& \textbf{0.931}
& \textbf{0.521}
& \underline{0.952}
& \textbf{0.990}
& \textbf{0.989}
& \textbf{0.983}
& \textbf{0.247} \\
\bottomrule
\end{tabular}
\end{threeparttable}
\end{table*}

\begin{table*}[t]
\centering
\small
\setlength{\tabcolsep}{2.8pt}
\renewcommand{\arraystretch}{1.08}
\begin{threeparttable}
\caption{
Ablation on negative sample construction and sample strategy.
We compare different negative sources and sampling strategies.
For each negative source, selective sampling is compared against uniform sampling:
\up denotes improvement, \down denotes degradation. Times(s) refers to the single-round sampling time. 
Best results are highlighted in \textbf{bold}, and second-best results are underlined. 
}
\label{abla_neg}
\begin{tabular}{p{1.2cm}cccccccc}
\toprule
\textbf{Negative}
& \textbf{Time (s)}
& \multicolumn{2}{c}{\textbf{Authenticity}}
& \multicolumn{5}{c}{\textbf{VBench}} \\
\cmidrule(lr){3-4} \cmidrule(lr){5-9}
& 
& \textbf{Forensic} $\uparrow$
& \textbf{Om-D} $\uparrow$
& \textbf{Background} $\uparrow$
& \textbf{Motion} $\uparrow$
& \textbf{Subject} $\uparrow$
& \textbf{Temporal} $\uparrow$
& \textbf{Overall} $\uparrow$ \\
\midrule

T2V
& 15
& 0.830 & 0.486 & 0.931 & 0.973 & 0.919 & 0.958 & 0.162 \\
\midrule

Frame
& 15
& 0.829 & 0.485 & 0.935 & 0.974 & 0.926 & 0.959 & 0.159 \\
\hspace{0.6em}+ conc.
& 19
& 0.844\,\up
& 0.512\,\up
& 0.931\,\down
& 0.975\,\up
& 0.927\,\up
& 0.960\,\up
& 0.160\,\up \\
\midrule

V2V
& 15
& 0.822 & 0.488 & 0.934 & 0.976 & 0.928 & 0.964 & 0.158 \\
\hspace{0.6em}+ conc.
& 19
& 0.837\,\up
& 0.511\,\up
& 0.940\,\up
& 0.978\,\up
& 0.931\,\up
& 0.966\,\up
& 0.157\,\down \\
\midrule

Noise
& 4
& 0.862
& 0.494
& 0.938
& 0.979
& 0.925
& 0.966
& 0.157 \\
\rowcolor{bestgray}
\hspace{0.6em}+ conc.
& 6
& \textbf{0.876}\,\up
& \textbf{0.524}\,\up
& \textbf{0.947}\,\up
& \textbf{0.989}\,\up
& \textbf{0.937}\,\up
& \textbf{0.961}\,\down
& \textbf{0.168}\,\up \\
\bottomrule
\end{tabular}
\end{threeparttable}
\end{table*}
\subsection{Experimental Setup}
\paragraph{Implementation details.}
We train on the training splits of MotionBench~\cite{hong2025motionbench} and WISA~\cite{wangwisa}, and evaluate on their held-out test prompts following~\cite{wudensedpo}. MotionBench is a motion-centric benchmark for evaluating temporal reasoning and motion realism in text-to-video generation. WISA is a physics-aware text-to-video dataset covering diverse physical laws and scenarios. Since WISA does not provide an official split for preference learning, we construct our own train/test partition. Together, they enable evaluation of both motion fidelity and physics-aware temporal consistency. In order to compare the online construction methods of multiple types of negative samples, we instantiate our method on top of Wan-VACE~\cite{jiang2025vace}, an all-in-one video foundation model that supports multiple conditional generation modes. To isolate the effect of preference optimization on motion generation, we restrict all experiments to the text-to-video (T2V) setting and optimize only the T2V generation pathway. We fine-tunes it using LoRA~\cite{hu2022lora}, targeting video generation at a resolution of 240×416, with learning rate of 5e-6, mixed-precision (bf16) training, gradient accumulation, and EMA. We set $\Delta = 0.1$ to perform regularization.

\paragraph{Compared methods.} We compare our method against four representative baselines that differ in preference pair construction and supervision strategies: the pre-trained baseline refers to the original Wan-VACE checkpoint without preference optimization; DPO offline applies standard offline DPO using pairwise preferences derived from two model-generated videos under the same prompt, with preference labels determined by the VideoReward~\cite{liuimproving}; DPO offline (gt) introduces a variant that treats the ground-truth video as the positive sample and the generated video as the negative one, offering oracle supervision; DenseDPO\cite{wudensedpo} provides denser supervision by generating both positive and negative samples under real-video guidance and assigning preference scores at the temporal segment level. These baselines cover reward-model supervision, oracle preference construction, and dense temporal preference assignment, and therefore provide a representative comparison set for evaluation.

\paragraph{Evaluation metrics.} We evaluate generated videos from two complementary perspectives: authenticity and motion quality. For authenticity, we report Forensic-Chat~\cite{lin2025seeing}, a forensic realism score assessing visual authenticity and synthetic artifact suppression, along with OmniAID-Dino~\cite{guo2025omniaid}, an authenticity metric based on DINO that measures semantic realism and naturalness. Regarding motion quality, we follow the VBench~\cite{huang2024vbench} protocol and report scores for background consistency, motion smoothness, subject consistency, temporal flickering and overall consistency. 

\subsection{Quantitative Results}
Table~\ref{main_exp} and Table~\ref{main_exp_wisa_normal} present the quantitative results on the MotionBench and Wisa test sets, respectively. Our method consistently outperforms all baselines across both authenticity and motion-related metrics, demonstrating its effectiveness in improving perceptual realism while preserving motion fidelity. Compared with DenseDPO, our method yields a substantial gain in authenticity ($0.819 / 0.475 \rightarrow 0.876 / 0.524$), indicating that directly leveraging real videos as supervision provides a more reliable training signal than reward-model-based dense supervision that scores individual temporal segments. 

To disentangle the contribution of real videos as positive samples from the improvement introduced by our analysis framework itself, we conduct DPO (Online,GT) in which the ground-truth video is used as the positive sample and the generated video as the negative sample for both methods . Under this setting, our method still improves Forensic from 0.830 to 0.876 and OmniAID-D from 0.486 to 0.524. These gains indicate that the performance improvement is not solely due to the introduction of real data, but also arises from our preference attribution strategy. In particular, these results suggest that deriving supervision from the model’s own denoising trajectory and temporal concentration is an effective strategy.

\subsection{Qualitative Results}
\begin{figure}
  \centering
  \includegraphics[width=\linewidth]{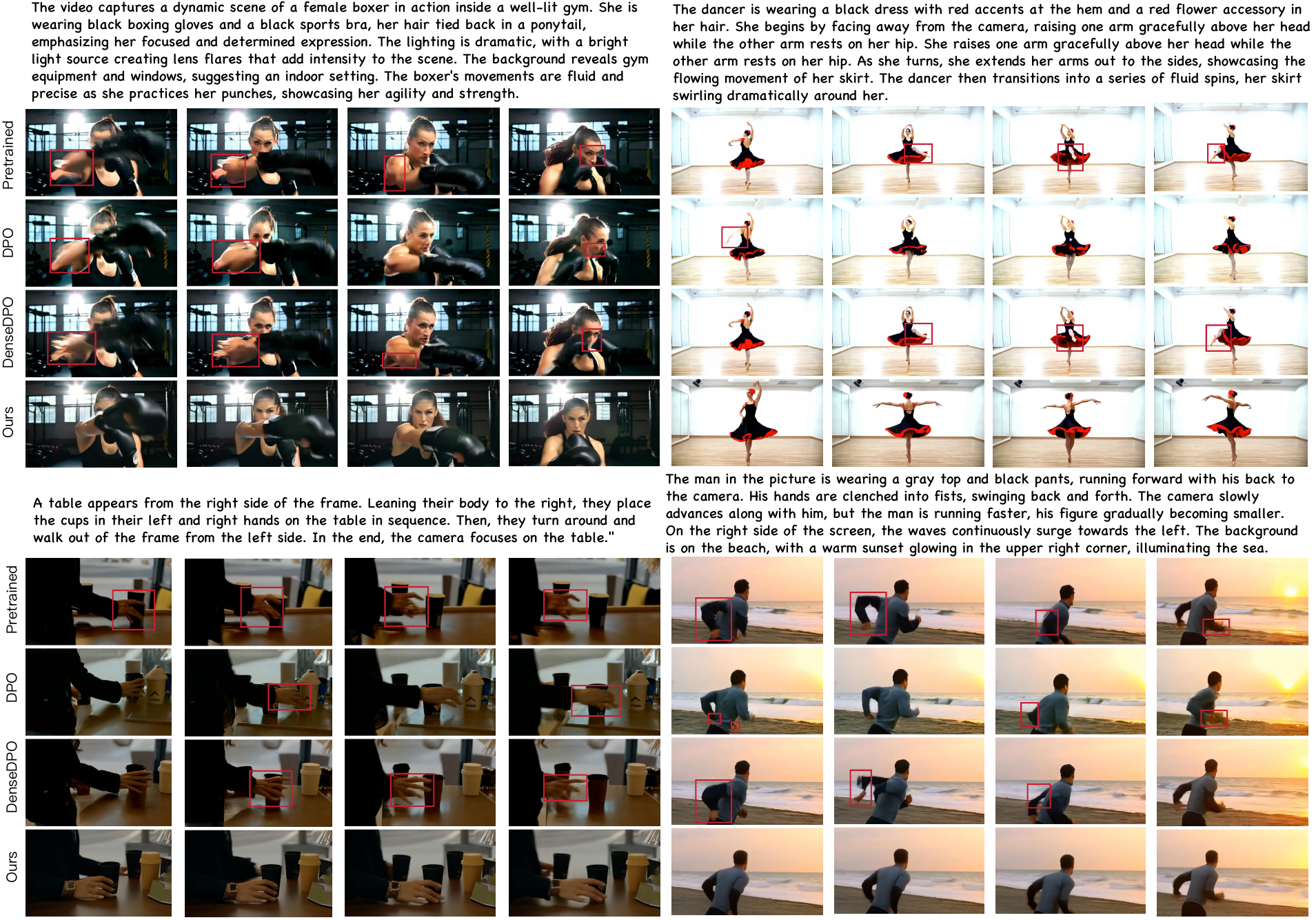}
  \caption{Qualitative Results. Compared to baselines, cIPO achieves better temporal coherence in challenging motion regions highlighted by the red boxes.}
  \label{vis}
  \vspace{-18pt}
\end{figure}

As shown in Fig.~\ref{vis}, compared with global preference optimization methods such as DPO and DenseDPO, cIPO produces markedly more temporally coherent videos across both fine-grained manipulation and authenticity. Specifically, in challenging motion regions highlighted by the red boxes, DPO and DenseDPO often exhibit noticeable temporal inconsistency, motion blur, and structural distortions, such as unstable hand-object interactions, inconsistent body poses during spinning motions, and unrealistic human motion trajectories. By contrast, cIPO maintains smoother temporal transitions, more stable object geometry, and more realistic motion dynamics throughout the video sequence. In contrast, cIPO leverages denoised-base implicit preference signals to identify failure-prone moments and concentrates optimization on high-error temporal windows, resulting in substantially improved authenticity and temporal stability.
\subsection{Ablation Studies}
\begin{figure}
  \centering
  \includegraphics[width=\linewidth]{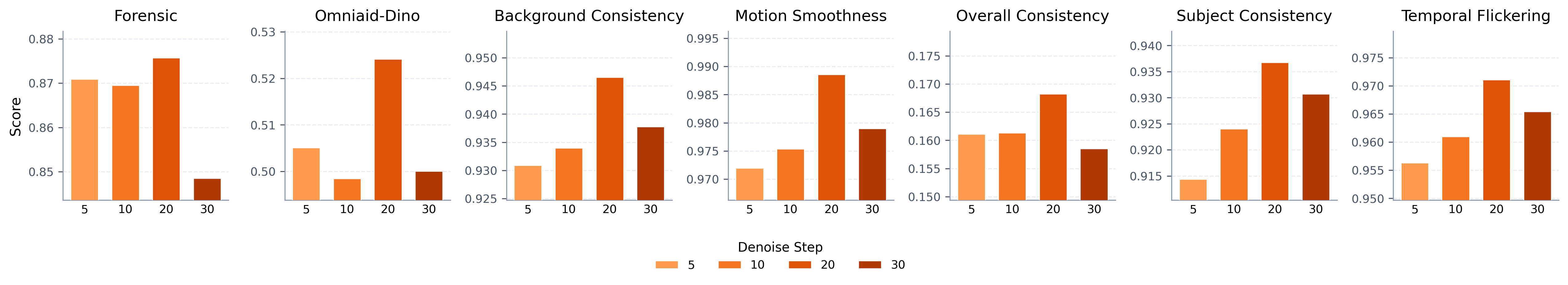}
  \caption{The impact of the change of the starting index $s$ of noise addition used in the negative sample sampling process on the video temporal quality and authenticity.}
  \label{abla_noise}
  \vspace{-12pt}
\end{figure}

We ablate the two core design choices in our method: how negative samples are constructed and how preference supervision is temporally allocated. Table~\ref{abla_neg} summarizes the results, while Fig.~\ref{abla_noise} further analyze noise starting step impact in our framework.

\textbf{Effect of negative sample from reconstruction rollouts.}
For negative sample construction, we leverage VACE to construct structurally aligned negatives through two controlled generation strategies: (1) first-frame-conditioned generation, in which negative samples share the identical initial latent as positive samples, and (2) video-to-video (V2V) generation, where the negative is generated under real-video guidance to remain close to the positive trajectory. For a fair comparison, all methods are tested under the same GPU-hour training budget. Similar to prior findings, these aligned pair construction strategies yield moderate improvements over unconstrained T2V negatives, particularly on motion-related metrics (as shown in Table.~\ref{abla_neg}). However, reconstruction-based noise negatives consistently outperform all alternative negative sources. This result supports our central design: negatives induced by diffusion perturbation are more informative than externally constructed samples because they better expose the model’s own inference-time failure modes. In addition, since denoise only requires a small number of sampling steps (about 5 to 20 steps), while other methods require complete sampling (50 steps), the training time and cost of the denoise method are lower.

\textbf{Effect of temporal concentration strategy.}
To validate the temporal concentration strategy, we apply it to all negative sample variants (excluding T2V due to substantial disparity between positive and negative samples that hinder direct comparison). As shown in Table.~\ref{abla_neg}, the consistent performance gains confirm its orthogonality with respect to the negative sampling source. This demonstrates the effectiveness and universality of the temporal concentration strategy.

\textbf{Noise starting step.} Fig.~\ref{abla_noise} examines the diffusion starting index sfor constructing reconstruction negatives. Results show that moderate noise levels work best: too little noise yields overly simple negatives, while excessive noise disrupts reconstruction and semantic consistency. This observation is consistent with the intuition that useful preference pairs should be difficult enough to expose rollout failure, but not so corrupted as to become semantically uninformative.
\subsection{Analysis}
\begin{figure}
  \centering
  \includegraphics[width=\linewidth]{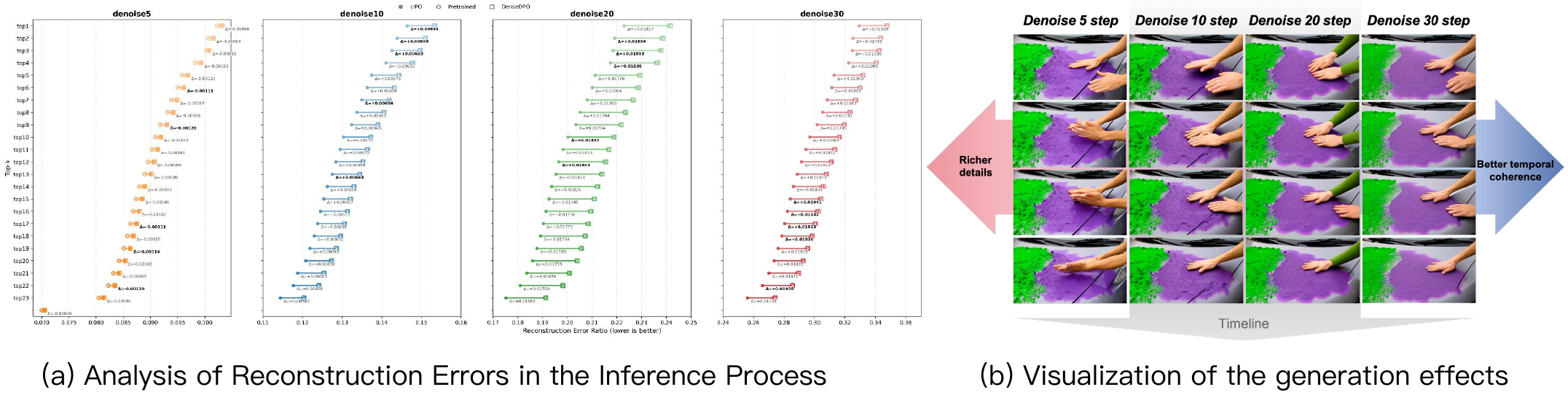}
  \caption{Analysis of the effect of different noise levels}
  \label{Analysis}
  \vspace{-15pt}
\end{figure}
\textbf{Reconstruction Errors in the Inference Process.} Fig.~\ref{Analysis}.a provides a comprehensive validation of the two core design choices in cIPO: implicit preference construction and temporally concentrated optimization. The x-axis shows the Reconstruction Error Ratio (lower is better), while the y-axis denotes the top-k highest-error temporal windows selected for preference optimization. The four subplots correspond to different forward noise levels, where larger denoise steps indicate stronger forward perturbation and thus a more challenging reconstruction task. Across all noise levels, cIPO consistently outperforms Pretrained and DenseDPO, demonstrating that constructing preference pairs from the original video and its denoised reconstruction provides a more effective supervision signal.
Specifically, under a moderate number of noise level (e.g., 10 and 20), as shown in Fig.~\ref{abla_noise}, the model achieves the best authenticity and motion quality trade-off. This is likely because most severely corrupted frames are corrected (as illustrated in Fig.~\ref{Analysis}.a, the most significant improvements are concentrated in the first four steps for the 10-noise and 20-noise settings), leading to substantial performance gains, which further validates the effectiveness of our temporal concentration strategy.

\textbf{Visualization of the generation effects of models.}  The generation results under different noise levels reflect the natural trade-off between authenticity and temporal consistency. As shown in Fig.~\ref{Analysis}.b, under weak noise, the reconstruction results can better preserve local textures and spatial details, but they are insufficient in exposing temporal artifacts, and local jitter and inter-frame flicker are still likely to occur. Under strong noise, although the model can learn smoother motion patterns and more stable cross-frame consistency, it often comes with detail blurring and texture loss. This phenomenon indicates that cIPO lies in constructing preference pairs using noise of appropriate level, enabling the model to achieve a better compromise between visual details and temporal consistency.

\section{Conclusions and Limitations}
In this work, we propose concentrated Implicit Preference Optimization (cIPO), a post-training framework for improving temporal coherence in text-to-video generation. By deriving preference signals directly from the model’s denoising process, cIPO avoids the need for costly human annotations or unstable external reward models, enabling efficient and rollout-consistent preference learning. Moreover, cIPO identifies temporally localized failure regions and concentrates optimization on high-error segments, leading to more precise training for short but perceptually critical artifacts. Experimental results show that cIPO consistently improves both authenticity and temporal consistency across multiple datasets, demonstrating the effectiveness of denoised-based preference with temporally focused optimization. However, cIPO still relies on reconstruction error as a proxy for perceptual quality, which may not fully capture high-level semantic preferences or human subjective judgments.

\clearpage
\renewcommand{\refname}{References}
\bibliographystyle{plain}   
\bibliography{references}   

@inproceedings{wudensedpo,
  title={DenseDPO: Fine-Grained Temporal Preference Optimization for Video Diffusion Models},
  author={Wu, Ziyi and Kag, Anil and Skorokhodov, Ivan and Menapace, Willi and Mirzaei, Ashkan and Gilitschenski, Igor and Tulyakov, Sergey and Siarohin, Aliaksandr},
  booktitle={The Thirty-ninth Annual Conference on Neural Information Processing Systems},
  year={2025}
}

@InProceedings{Yuan_2025_ICCV,
    author    = {Yuan, Zhihang and Xie, Rui and Shang, Yuzhang and Zhang, Hanling and Wang, Siyuan and Yan, Shengen and Dai, Guohao and Wang, Yu},
    title     = {DLFR-Gen: Diffusion-based Video Generation with Dynamic Latent Frame Rate},
    booktitle = {Proceedings of the IEEE/CVF International Conference on Computer Vision (ICCV)},
    month     = {October},
    year      = {2025},
    pages     = {16410-16419}
}

@article{yuan2025vgdfr,
  title={VGDFR: Diffusion-based Video Generation with Dynamic Latent Frame Rate},
  author={Yuan, Zhihang and Xie, Rui and Shang, Yuzhang and Zhang, Hanling and Wang, Siyuan and Yan, Shengen and Dai, Guohao and Wang, Yu},
  journal={arXiv preprint arXiv:2504.12259},
  year={2025}
}

@inproceedings{yuan2025dlfr,
  title={Dlfr-vae: Dynamic latent frame rate vae for video generation},
  author={Yuan, Zhihang and Wang, Siyuan and Shang, Yuzhang and Zhang, Hanling and Fang, Tongcheng and Xie, Rui and Yan, Shengen and Dai, Guohao and Wang, Yu},
  booktitle={Proceedings of the 33rd ACM International Conference on Multimedia},
  pages={10388--10397},
  year={2025}
}

@inproceedings{liuimproving,
  title={Improving Video Generation with Human Feedback},
  author={Liu, Jie and Liu, Gongye and Liang, Jiajun and Yuan, Ziyang and Liu, Xiaokun and Zheng, Mingwu and Wu, Xiele and Wang, Qiulin and Xia, Menghan and Wang, Xintao and others},
  booktitle={The Thirty-ninth Annual Conference on Neural Information Processing Systems},
  year={2025}
}

@article{wang2024lift,
  title={Lift: Leveraging human feedback for text-to-video model alignment},
  author={Wang, Yibin and Tan, Zhiyu and Wang, Junyan and Yang, Xiaomeng and Jin, Cheng and Li, Hao},
  journal={arXiv preprint arXiv:2412.04814},
  year={2024}
}

@article{wangunified,
  title={Unified Reward Model for Multimodal Understanding and Generation},
  author={Wang, Yibin and Zang, Yuhang and Li, Hao and Jin, Cheng and Wang, Jiaqi},  journal={arXiv preprint arXiv:2503.05236},
  year={2025}
}

@article{zhang2024onlinevpo,
  title={Onlinevpo: Align video diffusion model with online video-centric preference optimization},
  author={Zhang, Jiacheng and Wu, Jie and Chen, Weifeng and Ji, Yatai and Xiao, Xuefeng and Huang, Weilin and Han, Kai},
  journal={arXiv preprint arXiv:2412.15159},
  year={2024}
}

@article{cheng2025discriminator,
  title={Discriminator-free direct preference optimization for video diffusion},
  author={Cheng, Haoran and Dong, Qide and Peng, Liang and Sha, Zhizhou and Feng, Weiguo and Xie, Jinghui and Song, Zhao and Wen, Shilei and He, Xiaofei and Wu, Boxi},
  journal={arXiv preprint arXiv:2504.08542},
  year={2025}
}

@article{huang2026mind,
  title={Mind the Generative Details: Direct Localized Detail Preference Optimization for Video Diffusion Models},
  author={Huang, Zitong and Zhang, Kaidong and Ding, Yukang and Gao, Chao and Ding, Rui and Chen, Ying and Zuo, Wangmeng},
  journal={arXiv preprint arXiv:2601.04068},
  year={2026}
}

@article{cheng2025realdpo,
  title={RealDPO: Real or Not Real, that is the Preference},
  author={Cheng, Guo and Yang, Danni and Huang, Ziqi and Si, Jianlou and Si, Chenyang and Liu, Ziwei},
  journal={arXiv preprint arXiv:2510.14955},
  year={2025}
}

@inproceedings{jiang2025vace,
  title={Vace: All-in-one video creation and editing},
  author={Jiang, Zeyinzi and Han, Zhen and Mao, Chaojie and Zhang, Jingfeng and Pan, Yulin and Liu, Yu},
  booktitle={Proceedings of the IEEE/CVF International Conference on Computer Vision},
  pages={17191--17202},
  year={2025}
}

@article{lin2025seeing,
  title={Seeing before reasoning: A unified framework for generalizable and explainable fake image detection},
  author={Lin, Kaiqing and Yan, Zhiyuan and Chen, Ruoxin and Ye, Junyan and Zhang, Ke-Yue and Zhou, Yue and Jin, Peng and Li, Bin and Yao, Taiping and Ding, Shouhong},
  journal={arXiv preprint arXiv:2509.25502},
  year={2025}
}

@article{guo2025omniaid,
  title={OmniAID: Decoupling Semantic and Artifacts for Universal AI-Generated Image Detection in the Wild},
  author={Guo, Yuncheng and Ye, Junyan and Zhang, Chenjue and Kang, Hengrui and Fu, Haohuan and He, Conghui and Li, Weijia},
  journal={arXiv preprint arXiv:2511.08423},
  year={2025}
}

@inproceedings{huang2024vbench,
  title={Vbench: Comprehensive benchmark suite for video generative models},
  author={Huang, Ziqi and He, Yinan and Yu, Jiashuo and Zhang, Fan and Si, Chenyang and Jiang, Yuming and Zhang, Yuanhan and Wu, Tianxing and Jin, Qingyang and Chanpaisit, Nattapol and others},
  booktitle={Proceedings of the IEEE/CVF Conference on Computer Vision and Pattern Recognition},
  pages={21807--21818},
  year={2024}
}

@inproceedings{wangwisa,
  title={WISA: World simulator assistant for physics-aware text-to-video generation},
  author={Wang, Jing and Ma, Ao and Cao, Ke and Zheng, Jun and Feng, Jiasong and Zhang, Zhanjie and Pang, Wanyuan and Liang, Xiaodan},
  booktitle={The Thirty-ninth Annual Conference on Neural Information Processing Systems},
  year={2025}
}

@inproceedings{hong2025motionbench,
  title={Motionbench: Benchmarking and improving fine-grained video motion understanding for vision language models},
  author={Hong, Wenyi and Cheng, Yean and Yang, Zhuoyi and Wang, Weihan and Wang, Lefan and Gu, Xiaotao and Huang, Shiyu and Dong, Yuxiao and Tang, Jie},
  booktitle={Proceedings of the Computer Vision and Pattern Recognition Conference},
  pages={8450--8460},
  year={2025}
}

@article{pal2024smaug,
  title={Smaug: Fixing failure modes of preference optimisation with dpo-positive},
  author={Pal, Arka and Karkhanis, Deep and Dooley, Samuel and Roberts, Manley and Naidu, Siddartha and White, Colin},
  journal={arXiv preprint arXiv:2402.13228},
  year={2024}
}

@inproceedings{songscore,
  title={Score-Based Generative Modeling through Stochastic Differential Equations},
  author={Song, Yang and Sohl-Dickstein, Jascha and Kingma, Diederik P and Kumar, Abhishek and Ermon, Stefano and Poole, Ben},
  booktitle={International Conference on Learning Representations},
  year={2020}
}

@inproceedings{lipmanflow,
  title={Flow Matching for Generative Modeling},
  author={Lipman, Yaron and Chen, Ricky TQ and Ben-Hamu, Heli and Nickel, Maximilian and Le, Matthew},
  booktitle={The Eleventh International Conference on Learning Representations},
  year={2022}
}

@article{hu2022lora,
  title={Lora: Low-rank adaptation of large language models.},
  author={Hu, Edward J and Shen, Yelong and Wallis, Phillip and Allen-Zhu, Zeyuan and Li, Yuanzhi and Wang, Shean and Wang, Liang and Chen, Weizhu and others},
  journal={Iclr},
  volume={1},
  number={2},
  pages={3},
  year={2022}
}

\appendix


\section{Analysis of Gradient Degeneration with Near-Identical Pairs}
\label{sec:flowdpo_gradient_degenerate}
The design of winner-preservation penalty is primarily motivated by a key failure mode in Flow-DPO: when the preferred and dispreferred trajectories are similar, the probability of the preferred sample can actually decrease. In this section, we analyze the causes of this problem.

\paragraph{Setup}

Consider a flow model parameterized by $\theta$ with velocity predictor
$v_\theta(x_t,t)$.
For a trajectory sample $(x_t,v)$ at time $t$, the conditional likelihood is

\[
\log p_\theta(v \mid x_t,t)
=
-\frac{\beta_t}{2}
\|v-v_\theta(x_t,t)\|^2
+ C,
\]

where $\beta_t>0$ is a time-dependent weighting coefficient and $C$ is independent of $\theta$.

Given a preferred trajectory $w$ and a dispreferred trajectory $l$, Flow-DPO optimizes the objective

\[
\mathcal L_{\mathrm{FDPO}}
=
-\mathbb E
\left[
\log \sigma\left(
\Delta_\theta
\right)
\right],
\]

where

\[
\Delta_\theta
=
\log p_\theta(v^w \mid x_t^w,t)
-
\log p_\theta(v^l \mid x_t^l,t)
-
\left(
\log p_{\mathrm{ref}}(v^w \mid x_t^w,t)
-
\log p_{\mathrm{ref}}(v^l \mid x_t^l,t)
\right).
\]

Since the reference model is fixed, the optimization dynamics are governed by

\[
\Delta_\theta
=
-\frac{\beta_t}{2}
\left(
\|v^w-v_\theta(x_t^w,t)\|^2
-
\|v^l-v_\theta(x_t^l,t)\|^2
\right)
+ \mathrm{const}.
\]

We now formalize the causal relationship between gradient competition and the decrease of preferred likelihood in Flow-DPO.

Recall that the Flow-DPO update direction is

\[
\delta\theta
\propto
g_w-g_l,
\]

where

\[
g_w
=
\nabla_\theta
\log p_\theta(v^w\mid x_t^w,t),
\qquad
g_l
=
\nabla_\theta
\log p_\theta(v^l\mid x_t^l,t).
\]

To analyze how this update affects the preferred trajectory, we consider the first-order change of the preferred log-likelihood:

\[
\delta
\log p_\theta(v^w\mid x_t^w,t)
\approx
g_w^\top \delta\theta.
\]

Substituting the Flow-DPO update gives

\[
\delta
\log p_\theta(v^w\mid x_t^w,t)
\propto
g_w^\top(g_w-g_l).
\]

Expanding the inner product yields

\[
\delta
\log p_\theta(v^w\mid x_t^w,t)
\propto
\underbrace{\|g_w\|^2}_{\text{preferred enhancement}}
-
\underbrace{g_w^\top g_l}_{\text{dispreferred suppression}}.
\]

This decomposition reveals two competing forces:

\begin{itemize}
\item
The term $\|g_w\|^2$ corresponds to the standard likelihood-increasing effect for the preferred trajectory.

\item
The term $g_w^\top g_l$ arises from suppressing the dispreferred trajectory and measures how much this suppression interferes with the preferred update direction.
\end{itemize}

\paragraph{Similarity induces gradient alignment.}

When the preferred and dispreferred trajectories are temporally similar, i.e.,

\[
x_t^w \approx x_t^l,
\qquad
t_w \approx t_l,
\]

their network Jacobians become highly correlated:

\[
\nabla_\theta v_\theta(x_t^w,t)
\approx
\nabla_\theta v_\theta(x_t^l,t).
\]

Consequently, the corresponding likelihood gradients become aligned:

\[
g_w^\top g_l
\approx
\|g_w\|\,\|g_l\|.
\]

\paragraph{Why the dispreferred suppression term may dominate.}

After Flow Matching pretraining or supervised finetuning, the preferred trajectory is typically already well fitted:

\[
\|v^w-v_\theta(x_t^w,t)\|
\ll
\|v^l-v_\theta(x_t^l,t)\|.
\]

Using

\[
g
=
\nabla_\theta
\log p_\theta(v\mid x_t,t)
=
\beta_t
(v-v_\theta(x_t,t))^\top
\nabla_\theta v_\theta(x_t,t),
\]

the gradient magnitude approximately scales with the residual norm:

\[
\|g\|
\propto
\|v-v_\theta(x_t,t)\|.
\]

Therefore,

\[
\|g_l\|>\|g_w\|.
\]

Combined with gradient alignment, this implies

\[
g_w^\top g_l
>
\|g_w\|^2.
\]

Hence,

\[
\delta
\log p_\theta(v^w\mid x_t^w,t)
<
0.
\]

Equivalently,

\[
p_\theta(v^w\mid x_t^w,t)
\]

decreases after the Flow-DPO update.

\paragraph{Interpretation.}

The key issue is that Flow-DPO optimizes a \emph{relative preference objective} rather than directly maximizing the absolute likelihood of preferred trajectories. Under strong overlap, suppressing the dispreferred trajectory requires parameter updates that are highly aligned with the preferred trajectory gradients. When the dispreferred gradient magnitude is larger, the suppression effect dominates the enhancement effect, causing the preferred likelihood to decrease despite optimizing a preference objective. In practice, this often manifests as spurious high-frequency textures or over-smoothed motion patterns. To address this limitation, we add a \emph{winner-preservation penalty}, which is activated when the current model underperforms the reference model on the preferred clean segment.

\paragraph{Experiment.}
\begin{wrapfigure}{r}{0.5\textwidth}
\vspace{-12pt} 
  \centering
  \includegraphics[width=\linewidth]{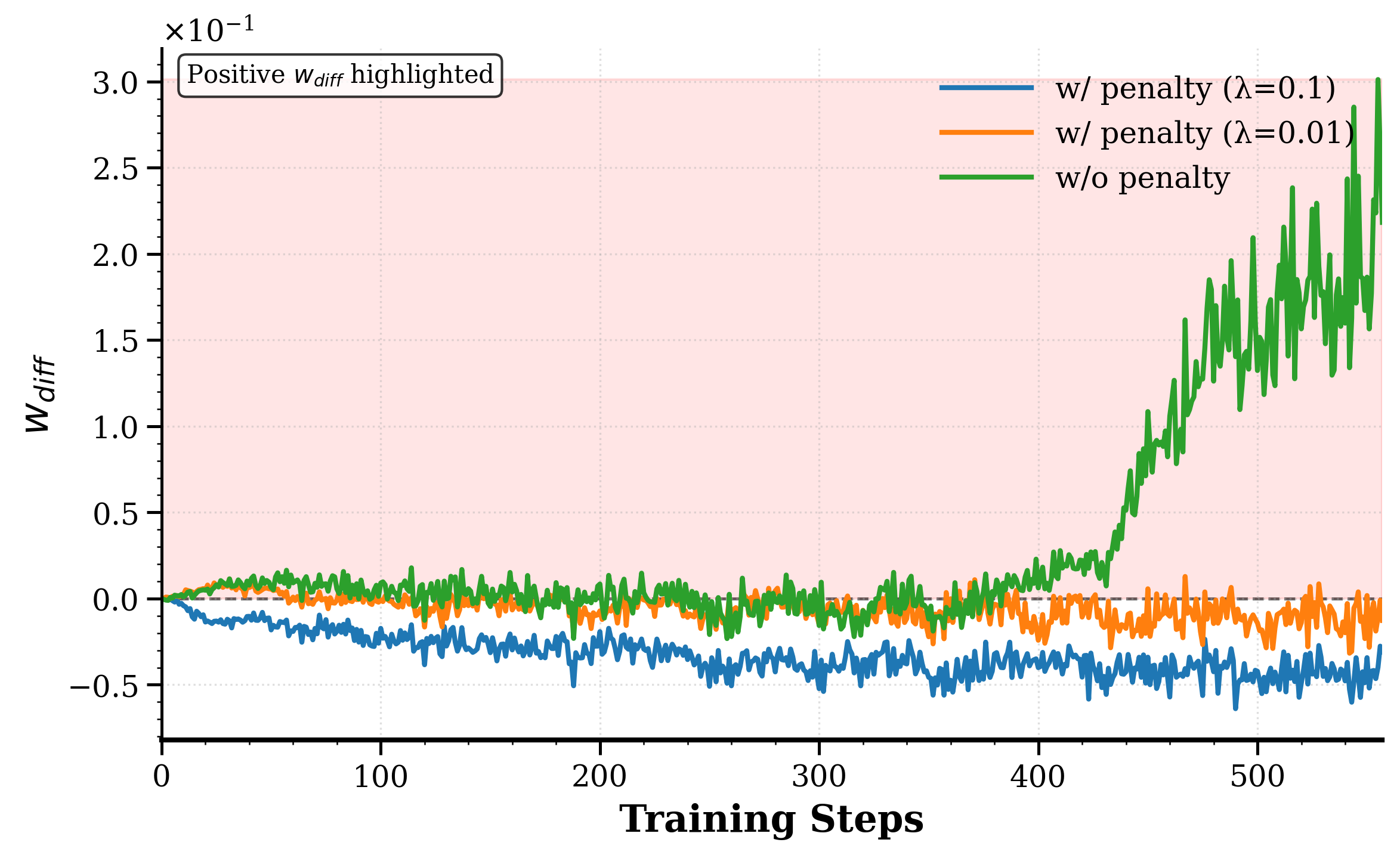}
  \caption{Variation of $w_{diff}$ during the training process with/without the penalty term.}
  \label{w_diff}
\vspace{-12pt} 
\end{wrapfigure}
In flow matching, the model learns a conditional velocity field \(v_\theta(x_t,c,t)\) by minimizing the regression objective
\[
\mathcal{L}_{\mathrm{FM}}
=
\mathbb{E}\big[\|v_\theta(x_t,c,t)-u_t\|^2\big],
\]
where \(u_t\) denotes the target velocity. Under the standard Gaussian regression interpretation,
\[
p_\theta(u_t \mid x_t,c,t)
\propto
\exp\!\left(
-\frac12
\|v_\theta-u_t\|^2
\right),
\]
which implies
\[
\log p_\theta(u_t \mid x_t,c,t)
=
-\frac12
\|v_\theta-u_t\|^2
+ \mathrm{const}.
\]
Therefore, for a preferred sample \(x_w\),
\[
w_{\mathrm{diff}}
=
\texttt{model\_win\_err}
-
\texttt{ref\_win\_err}
\]
corresponds to the relative log-likelihood change between the current model and the reference model:
\[
\log \frac{p_\theta(x_w)}{p_{\mathrm{ref}}(x_w)}
\propto
-\frac12\, w_{\mathrm{diff}}.
\]
Consequently, a larger positive \(w_{\mathrm{diff}}\) indicates that the current model assigns a lower probability to the preferred sample compared to the reference model.

As shown in Fig.~\ref{w_diff}, without the penalty term, \(w_{\mathrm{diff}}\) increases significantly during the later stage of training, indicating that the model progressively reduces the likelihood of preferred samples. This reveals a preference degeneration phenomenon where optimization unintentionally suppresses positive samples themselves. In contrast, with the proposed penalty, \(w_{\mathrm{diff}}\) remains close to zero or negative, demonstrating that the preferred-sample probability is preserved or improved throughout training. These results show that the penalty stabilizes preference optimization by preventing the collapse of preferred-sample likelihood.


\section{Human evaluation}
We conducted a human evaluation to compare four models: \textit{Pretrained}, \textit{DenseDPO}, \textit{DPO}, and \textit{cIPO}. For each sample, annotators were asked to select the best model output along six criteria: \textit{human structure}, \textit{motion dynamics}, \textit{scene structure}, \textit{object relations}, \textit{natural physics}, and \textit{temporal consistency}. The spreadsheet records the majority-vote winner for each sample--criterion pair. To obtain a dimension-level ranking, we counted how many samples were won by each model under the majority vote for that criterion. The resulting comparison is therefore based on majority-vote wins across samples rather than mean opinion scores. As summarized in Table~\ref{tab:human-eval-breakdown}, \textit{cIPO} is the strongest model on all six criteria, consistently achieving the highest majority-vote share for every dimension. The advantage of \textit{cIPO} is especially pronounced on \textit{natural physics} (72.4\%) and \textit{object relations} / \textit{temporal consistency} (69.0\% each), while the other three models remain substantially behind on every dimension.

\begin{table}[t]
\centering
\begin{minipage}{0.75\textwidth}
\centering
\caption{Percentages are computed within each dimension}
\label{tab:human-eval-breakdown}
\begin{tabular}{lcccc}
\toprule
Criterion & Pretrained & DenseDPO & DPO & cIPO \\
\midrule
Human Structure & 11.1\%& 11.1\%& 14.8\%& 63.0\% \\
Motion Dynamics & 6.9\%& 6.9\%& 20.7\%& 65.5\% \\
Scene Structure & 13.8\%& 10.3\%& 17.2\%& 58.6\% \\
Object Relations & 10.3\%& 6.9\%& 13.8\%& 69.0\% \\
Natural Physics & 10.3\%& 3.4\%& 13.8\%& 72.4\% \\
Temporal Consistency & 10.3\%& 10.3\%& 10.3\%& 69.0\% \\
\bottomrule
\end{tabular}
\end{minipage}
\end{table}

\section{Pseudocode for Rollout}
\label{app_rollout}
The pseudocode~\ref{Rollout} shows the specific construction method of negative samples in Implicit Preference Construction from Reconstruction Rollouts.

\begin{algorithm}
\caption{Rollout of a Noisy Latent Trajectory}
\label{Rollout}
\begin{algorithmic}[1]
\REQUIRE $\mathbf{z}_0$, scheduler $\mathcal{S}$, denoiser $\mathcal{M}$, 
prompt embedding $\mathbf{c}$, negative prompt $\mathbf{c}^{-}$,
steps $N$, guidance $w$, start step $s$
\ENSURE Denoised latent $\hat{\mathbf{z}}_0$

\STATE $\{t_k\}_{k=0}^{N-1} \leftarrow \mathcal{S}.\texttt{set\_timesteps}(N)$
\STATE $\{t_k\}_{k=s}^{N-1} \gets$ active steps
\STATE $\boldsymbol{\epsilon} \sim \mathcal{N}(\mathbf{0}, \mathbf{I})$
\STATE $\mathbf{z}_{t_s} \gets \mathcal{S}.\texttt{add\_noise}(\mathbf{z}_0, \boldsymbol{\epsilon}, t_s)$

\FOR{$t_k \in \{t_k\}_{k=s}^{N-1}$}
    \STATE $\hat{\boldsymbol{\epsilon}}_{\mathrm{cond}} \gets \mathcal{M}(\mathbf{z}_{t_k}, t_k, \mathbf{c})$
    
    \IF{$w > 1$ and $\mathbf{c}^{-}$ provided}
        \STATE $\hat{\boldsymbol{\epsilon}}_{\mathrm{uncond}} \gets \mathcal{M}(\mathbf{z}_{t_k}, t_k, \mathbf{c}^{-})$
        \STATE $\hat{\boldsymbol{\epsilon}} \gets
               \hat{\boldsymbol{\epsilon}}_{\mathrm{uncond}} + w\bigl(
               \hat{\boldsymbol{\epsilon}}_{\mathrm{cond}} - \hat{\boldsymbol{\epsilon}}_{\mathrm{uncond}}\bigr)$
    \ELSE
        \STATE $\hat{\boldsymbol{\epsilon}} \gets \hat{\boldsymbol{\epsilon}}_{\mathrm{cond}}$
    \ENDIF
    
    \STATE $\mathbf{z}_{t_{k+1}} \gets \mathcal{S}.\texttt{step}(\hat{\boldsymbol{\epsilon}}, t_k, \mathbf{z}_{t_k})$
\ENDFOR

\STATE $\hat{\mathbf{z}}_0 \gets \mathbf{z}_{t_N}$
\RETURN $\hat{\mathbf{z}}_0$
\end{algorithmic}
\end{algorithm}

\end{document}